\documentclass[10pt,twocolumn,letterpaper]{article}
\usepackage{3dv}
\usepackage{times}
\usepackage{epsfig}
\usepackage{graphicx}
\usepackage{amsmath}
\usepackage{amssymb}
\usepackage{color,soul}
\usepackage{subcaption}
\usepackage{cite}
\usepackage[mathscr]{euscript}
\newcommand{\norm}[1]{\left\lVert#1\right\rVert}

\newcommand\sm[1]{\textcolor{blue}{[SM: #1]}} 
\newcommand\eat[1]{}

\usepackage[pagebackref=true,breaklinks=true,letterpaper=true,colorlinks,bookmarks=false]{hyperref}
\hypersetup{
        colorlinks=true,
        citecolor=blue,
        urlcolor=red,
}

\threedvfinalcopy 

\def\threedvPaperID{131} 

\ifthreedvfinal\fi
\begin{document}

\title{Learning Point Embeddings from Shape Repositories for Few-Shot Segmentation}
\author{Gopal  Sharma \quad Evangelos Kalogerakis \quad Subhransu Maji \\
University of Massachusetts, Amherst\\
{\tt\small \{gopalsharma,kalo,smaji\}@cs.umass.edu}
}
\maketitle

\begin{abstract}
  User generated 3D shapes in online repositories contain rich information about
  surfaces, primitives, and their geometric relations, often arranged in a
  hierarchy. 
  We present a framework for learning representations of 3D shapes
  that reflect the information present in this meta data and show that it leads
  to improved generalization for semantic segmentation tasks. 
  Our approach is a
  point embedding network that generates a vectorial representation of the 3D
  points such that it reflects the grouping hierarchy and tag data. 
  The main
  challenge is that the data is noisy and highly variable. 
  To this end, we
  present a tree-aware metric-learning approach and
  demonstrate that such learned embeddings offer excellent transfer to semantic
  segmentation tasks, especially when training data is limited. Our approach
  reduces the relative error by $10.2\%$ with $8$ training examples, by
  $11.72\%$ with $120$ training examples on the ShapeNet semantic segmentation
  benchmark, in comparison to the network trained from scratch. 
  By utilizing tag data
  the relative error is reduced by $12.8\%$ with $8$ training examples, in
  comparison to the network trained from scratch. These improvements come at no
  additional labeling cost as the meta data is freely available.
\end{abstract}
\section{Introduction}
\begin{figure*}[ht!]
\centering
\includegraphics[width=\linewidth]{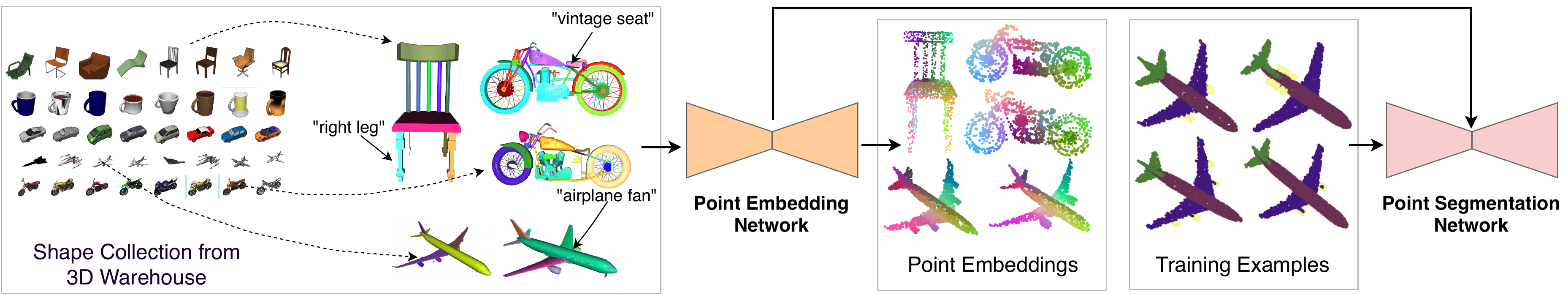}
\caption{\label{fig:teaser}
  \textbf{Overview of our approach.} Shape collections in 3D shape
  repositories contain metadata such as polygon groupings and tags
  assigned to parts. 
  These parts and tags assigned to them are highly variable, even within the same category. 
  We use the shapes and metadata to train a point embedding network that
  maps each point into a fixed dimensional vector (see Section~\ref{sec:method}
  and Figure~\ref{fig:architecture} for details.)
  The embeddings for a few shapes are 
  visualized as color channels using t-SNE mapping, where similar
  colors indicate correspondence across shapes.
  The learned parameters when used to initialize 
  a point segmentation network leads to improved
  performance when few training examples are available. \emph{(Please zoom in for details.)}}
\end{figure*}

\eat{
\begin{figure}[!htbp]
  \centering
  \label{fig:inconsistency}
  \includegraphics[width=0.8\linewidth]{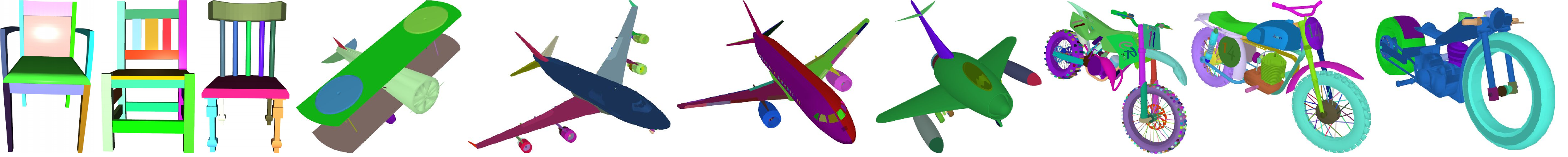}
  \caption{\textbf{Inconsiteny in mesh segmentation.} Segments generated by
    different designers tend to have huge variance, even for similar shapes,
    which gives different hierarchical arrangement of the shape-parts. This
    makes it hard to learn consistent segmentation across all shapes.}
\end{figure}
}

The ability to decompose a 3D shape into semantic parts can enable
applications from shape retrieval in online repositories, to robotic manipulation and shape generation.
Yet, automatic techniques for shape segmentation are limited by the
ability to collect labeled training data, which is often expensive or
time consuming.
Unlike images, online repositories of user-generated 3D shapes, such
as the 3D Warehouse repository~\cite{TrimbleWarehouse}, contain
rich metadata associated with each shape.
These include information about geometric primitives (e.g., polygons in 3D
meshes) organized in groups, often arranged in a hierarchy, as well as color,
material and tags assigned to them. 
This information reflects the modeling decisions of the designer are
likely correlated with high-level semantics.

\begin{figure*}[ht]
\centering
\includegraphics[width=\linewidth]{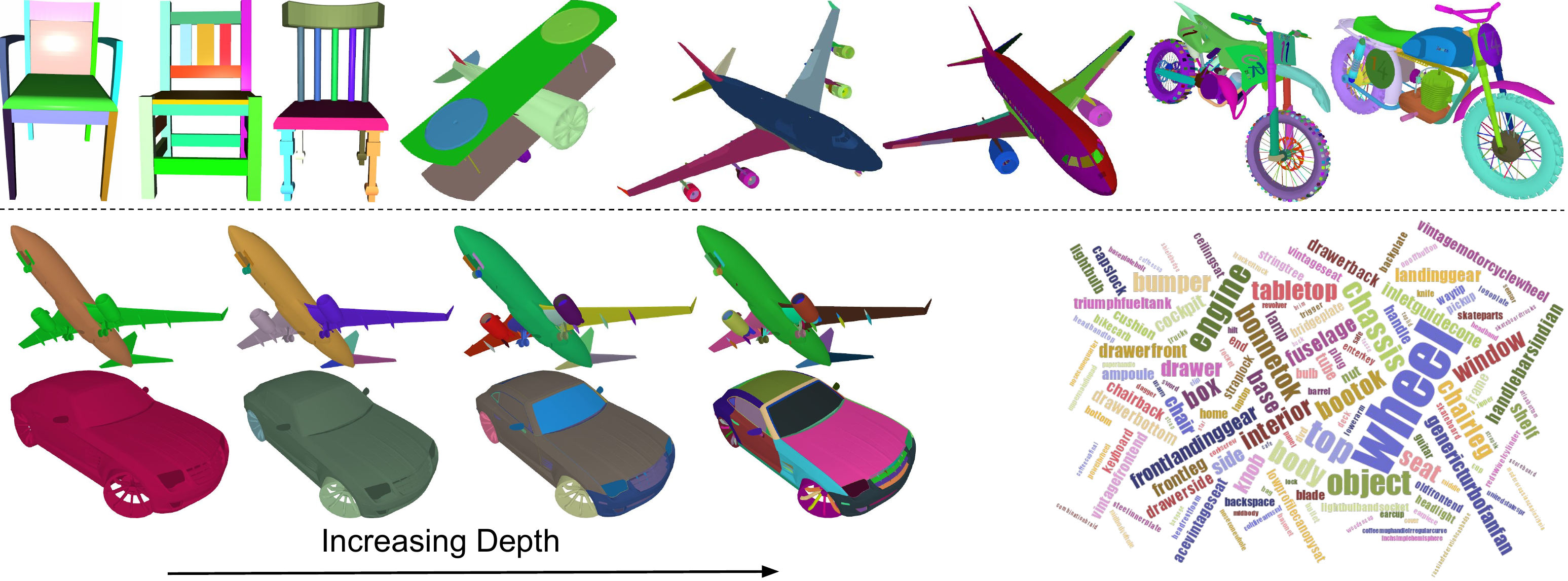}
\caption{(Top row) \textbf{Example shapes from our dataset.} Our dataset consist
  of shapes segmented into parts without any semantic information. Notice that
  shapes of same category can be segmented differently from each other. Here
  different color represents different leaf node in the part-hierarchy. (Bottom
  left) \textbf{Parts at different depths of the hierarchy} for an airplane and
  a car. Increasing the depth increases the number and granularity of parts.
  (Bottom right) \textbf{A word cloud} of raw tags collected from our dataset.
  The font size is proportional to the square root of frequency of the dataset.}
\label{fig:mesh-samples}
\end{figure*}

Despite its abundance, the use of metadata for learning shape
representations has been relatively unexplored in the literature.
One barrier is the high degree of its variability. 
These models were created by designers with a diverse set of goals and
with a wide range of expertise. 
As a result the groups and hierarchies over parts of a shape that reflect the modeling steps
taken by the designer are highly variable: two similar shapes can have
significantly different number of
parts as well as the number of levels in the part hierarchy. 
Moreover, the tags are rarely assigned to parts and are often
arbitrarily named. Figures~\ref{fig:teaser} and \ref{fig:mesh-samples} illustrate this variability.

Our work systematically addresses these challenges and presents an
approach to exploit the information present in the metadata to
improve the performance of a state-of-the-art 3D semantic segmentation
model.
Our approach, illustrated in Figure~\ref{fig:teaser}, consists of a deep
network that maps each
point in a 3D shape to a fixed dimensional
embedding. 
The network is trained in a way such that the embedding
reflects the user-provided hierarchy and tags.
We propose a robust tree-aware metric to supervise the point embedding
network that offers better
generalization to semantic segmentation tasks over a baseline scheme that is
tree-agnostic (only considers the leaf-level groupings). The point embedding
network trained on hierarchies also improves over models trained on shape
reconstruction tasks that leverage the 3D shape geometry but not their
metadata. Finally, when tags are available we show that the embeddings can be 
fine-tuned leading to further improvements in performance.

On the ShapeNet semantic segmentation dataset, an
embedding network pre-trained on hierarchy metadata outperforms a network
trained from scratch by reducing relative error by $10.2\%$ across 16
categories, when trained on $8$ shapes per category. 
Similarly, when only a small fraction of points (20 points)
per shape are labeled, the relative reduction in error is $4.9\%$.
Furthermore, on 5 categories which have sufficient tags, using both
the hierarchy and tags reduces error further by $12.8\%$ points
relative to the randomly initialized network, when trained on $8$ shapes per category. 
Our visualizations indicate that the trained networks
implicitly learn correspondences across shapes. 

\section{Related Work}

Our work builds on the advances in deep learning architectures for point-based,
or local, shape representations and metric learning approaches to guide
representation learning. We briefly review relevant work in these areas.

\paragraph{Supervised learning of local shape descriptors.} Several
architectures have been proposed to output local representations, or
descriptors, for 3D shape points or patches. The architectures can be broadly
categorized according to the type of raw 3D\ shape representation they consume.
Volumetric methods learn local patch representations by processing voxel
neighborhoods either in uniform \cite{maturana2015voxnets} or adaptively
subdivided grids
\cite{riegler2017octnet,klokov2017escape,wang2017ocnn,Wang:2018:AOP}. View or
multi-view approaches learning local image-based representations by processing
local 2D\ shape projections \cite{Huang:2017:LMVCNN,Tat2018}, which can be
mapped back onto the 3D shape \cite{kalogerakis2017shapepfcn}. Finally, a large
number of architectures have been recently proposed for processing raw point
clouds. PointNet and PointNet++ are transforming individual point coordinates
and optionally normals through MLPs and then performing permutation-invariant
pooling operations in local neighborhhoods
\cite{qi2017pointnet,qi2017pointnetpp}.

All the above-mentioned deep architectures are trained in a fully supervised
manner using significant amound of labeled data. Although for some specific
classes, like human bodies, these annotations can be easily obtained through
template-based matching or synthetically generated shapes
\cite{Allen:2003:SHB,Anguelov:2005:SSC,Bogo:CVPR:2014}, for the vast majorities
of shapes in online repositories, gathering such annotations often requires
laborious user interaction \cite{mo2018partnet,Yi:2016:SAF}. Active learning
methods have also been proposed to decrease the workload, but still rely on
expensive crowdsourcing \cite{Yi:2016:SAF}.\eat{Graph-based approaches learn
  local representations by treating the input shape as a connected graph
  \cite{masci2015geodesic}, often analyzing it in the spectral domain
  \cite{Boscaini:2015,Boscaini2016,Monti2017,Yi2017,bronstein2017geometric} or
  in a parameterized 2D domain \cite{Ezuz2017,Maron2017CNN}.} \eat{Other methods
  extend the convolution operation for point clouds
  \cite{hermosilla2018mccnn,PointCNN,Atzmon:2018:PCN}, or perform convolution in
  a high-dimensional lattice where the input points are embedded
  \cite{su18splatnet}.}

\paragraph{Weak supervision for learning shape descriptors.} A few methods
\cite{Muralikrishnan18,cosegnetChenyang} have been recently proposed to avoid
expensive point-based annotations. Muralikrishnan et al. \cite{Muralikrishnan18}
extracts point-wise representations by training an architecture designed to
predict shape-level tags (e.g., armrest chair) by first predicting intermediate
shape segmentations. Instead of using weak supervision in the form of
shape-level tags, we use unlabeled part hierarchies available in massive online
repositories and tags for parts (not whole shapes) when such are available. Yi
et al. \cite{Yi:2017:LHS} embeds pre-segmented parts in descriptor space by
jointly learning a metric for clustering parts, assigning tags to them, and
building a consistent part hierarchy. In our case, our architecture learns
point-wise descriptors and also relaxes the requirement of inferring consistent
hierarchies, which might be hard to estimate for shape families with significant
structural variability. Non-rigid geometric alignment has been used as a form of
weak and noisy supervision by extracting approximate local shape correspondences
between pairs of shapes of similar structure \cite{Huang:2013:FSL} or by
deforming part templates \cite{Huang:2015:dbm}. However, global shape alignment
can fail for shapes with different structure, while part-based alignment
requires corresponding parts or definition of part templates in the first place.
In a concurrent work, given a collection of shapes from a single category, Chen
\etal \cite{BAENet} proposed a branched autoencoder that discovers coarse
segmentations of shapes by predicting implicit fields for each part. Their
network is trained with a few manually selected labeled shapes in a few-shot
semantic segmentation setting. Our method instead utilizes part hierarchies and
metadata as weak supervisory signal. We also randomly select labeled sets for
our few-shot experiments. In general, our method is complementary to all the
above-mentioned weak supervision methods. Our weak signal in the form of
unlabeled part hierarchies and part tags can be used in conjunction with
geometric alignment, consistent hierarchies, or shape-level tags, whenever such
are possible to obtain.


\paragraph{Triplet-based metric learning.}
Our approach learns a metric embedding over points that reflects the hierarchies
in 3D shape collections. Metric learning has a rich literature with a diverse
applications and techniques. A popular approach is to supervise the learning
with ``triplets'' of the form $(a, b, c)$ to denote that \emph{``a is more similar to
b than c''}.
This can be written as $d(a, b) \leq d(a,c)$ where the
$d(a,b)$ denotes the distance between $a$ and $b$. 
The distance itself could be computed as the Euclidean distance
in some embedding space, i.e., $d(a,b) = ||\phi(a) -
\phi(b)||_2$, possibly computed with a deep network.
Within this
framework, techniques to sample triplets remains an active area of research. These
include techniques such as hard-negative mining \cite{Hermans2017InDO},
semi-hard negative mining \cite{SchroffKP15} and distance weighted sampling
\cite{Wu} to bias the sampling of triplets.

\section{Mining Metadata from Shape Repositories}
\label{sec:dataset}
We first describe the source of our part hierarchy dataset that we use for
training our embedding network. Then we describe the metadata (tags) present in the
3d models and how we extract this information into a consistent dataset.
\vspace{-0.1cm}
\paragraph{Part hierarchies.}
Several 3D modeling tools, such as SketchUp, Maya, 3DS Max to name a few, allow
users to model shapes, and scenes, in general, as a collection of geometric
entities (e.g., polygons) organized into groups. The groups can be nested and
organized in hierarchies. In our part hierarchy dataset, we endeavor to extract
these hieararchies. The shapes in our dataset are a subset of Shapenet Core dataset,
where we focus on $16$ categories from Shapenet part-segmentation dataset
\cite{Yi:2016:SAF} to allow systematic benchmarking and comparison with prior
work. Note that the $16$ categories semantic segmentation dataset contains
$~16.6k$ shapes, whereas $16$ categories in Shapenet Core dataset contains
$~28k$ shapes. We first retrieved the original files for shapes in Shapenet Core
dataset provided by 3d warehouse, which are stored in the popular ``COLLADA''
format \cite{wiki:collada}. These files represent 3D models in a hierarchical
tree structure. Leaf nodes represent shape geometry, and internal nodes
represent groups of geometric primitives, or nested groups. Samples from our
dataset are visualized in the Figure \ref{fig:mesh-samples}. Number of parts in
which a shape is segmented depends on the part-hierarchy as visualized in the
Figure \ref{fig:mesh-samples} (bottom left). Models with too few part
segmentation (less than $2$) or too many (more than $500$) are discarded. This
gives us a total of $20776$ 3D\ models having part group information, with each
model having at least one level of part grouping. We further segment the dataset
into train ($15625$), validation ($3113$) and test ($2038$) splits. We ensure
that the shapes in test split of semantic part-segmentation dataset
\cite{Yi:2016:SAF} are not included in the train split of our part hierarchy
dataset.
\eat{For example, in an car model, a designer might create
  various geometric primitives that correspond to different, small parts of its
  frame and wheels, then group those to larger parts, such as the entire frame
  and wheels. Grouping is up to the designer. The designer might design a 3D
  model as a collection of nested part hierarchies, or use just a flat group of
  parts, or even use no groups at all (i.e., model the shape as a continuous,
  connected surface). \sm{This paragraph is repetative from earlier.}}


\begin{table}[t!]
  \begin{tabular}{c|c|c}
Category  &  Shapes with part tagged & Avg points tagged\\
\hline
Motorcycle & $110$           & $11.3\%$\\
Airplane   & $806$           & $5.0\%$\\
Table      & $392$           & $45.7\%$\\
Chair      & $326$           & $38.7\%$\\
Car        & $600$           & $20.0\%$\\
\hline
\end{tabular}
\vskip -3mm
\caption{\textbf{Dataset with tags.} Number of shapes with at least one tagged
  parts, and average percentage of points tagged in these shapes in 5
  categories.}
\label{table:dataset-tags}
\end{table}
\paragraph{Tag extraction.} Modeling tools  allow users to explicitly give
tags to parts, which are stored in their corresponding file format. Obviously,
not all designers enter tags for their designed parts. Out of all the models that
include part group information in our dataset, we observed that only $10.7\%$ of
the shapes had meaningful tags for at least one part (i.e., tags are sparse).
Usually, these tags are not consistent, e.g., a tag for a wheel part in a car can be
``wheel\_mesh''. 
To make things worse, few tags have high frequency e.g., one
may encounter wheel, chassis, windows (or synthetics of those) frequently as
tags, while most of them are rare, or even be non-informative for part types
e.g., ``geometry123''.

To extract meaningful tags, we selected the $10$ most frequent tags encountered
as strings, or sub-strings stored in the nodes for each shape category.
We also merge synonyms into one tag to reduce number of tags in the final set.
For every tag, we find the corresponding geometry nodes and then we label the
points sampled from these nodes with the tag. We found that only $5$ out of $16$
categories have a ``sufficient'' number of tagged points ($>1\%$ of the original
surface points). By ``sufficient'', we mean that below this threshold, tags are
becoming so sparse in a category that result in negligible improvements. Table
\ref{table:dataset-tags} shows the distribution of tags in these $5$
categories.

\paragraph{Geometric postprocessing.} 
We finally aligned the shapes using ICP so that their orientation agrees with the
canonical orientation provided for the same shapes in ShapeNet. To process the
shapes through our point-based architecture, we uniformly sampled 10K points on
their surface. Further details about these steps are provided in the
supplementary material.


\section{Method} \label{sec:method}
\begin{figure*}[!ht]
\label{fig:architecture}
\centering
\includegraphics[width=\textwidth]{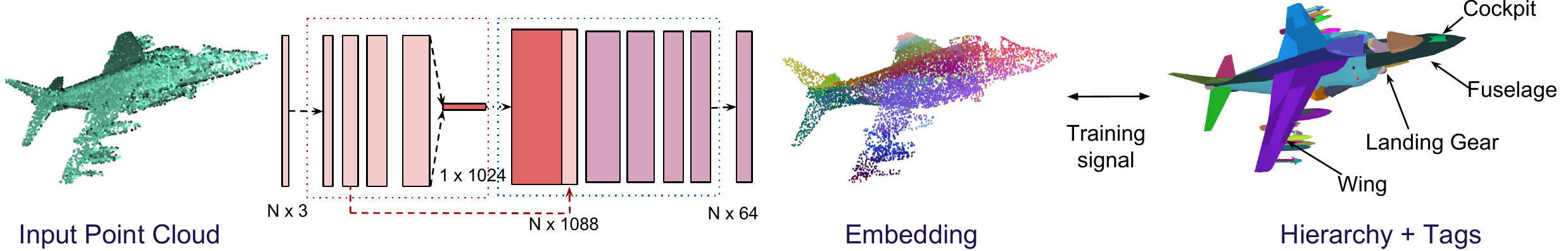}
\caption{\textbf{Architecture of the Point Embedding Network (PEN)}. The network
  takes as input a point cloud and outputs a fixed dimensional embedding for
  each point, visualized here using t-SNE. These embeddings are learned using
  metric learning that utilizes part-hierarchy. Furthermore, embedding can be
  improved by supervising network using sparsely tagged point cloud from a small
  subset of our dataset (refer Table \ref{table:dataset-tags}). Tags are pointed
by arrows.}
\label{fig:architecture}
\end{figure*}

Our Point Embedding Network (PEN) takes as input a shape in the form of a point
cloud set, $X = \{\mathbf{x}_i\}_{i=1}^N $, where $\mathbf{x}$ represents the 3D
coordinates of each point. Our network learns to map each input shape point
$\mathbf{x}$ to an embedding $\phi_{\mathbf{w}}(\mathbf{x}) \in {\cal R}^d$
based on learned network parameters $\mathbf{w}$. The architecture is
illustrated in Figure~\ref{fig:architecture}. PEN first incorporates a PointNet
module \cite{qi2017pointnet}: the points in the input shape are individually
encoded into vectorial representations through MLPs, then the resulting
point-wise representations are aggregated through max pooling to form a global
shape representation. The representation is invariant to the order of the points
in the input point set. At the next stage, the learned point-wise
representations are concatenated with the global shape representation, and are
further transformed through fully-connected layers and ReLUs. In this manner,
the point embeddings reflect both local and global shape information.

We used PointNet as a module to extract the initial point-wise and global shape
representation mainly due to its efficiency. In general, other point-based
modules, or even volumetric
\cite{maturana2015voxnets,wang2017ocnn,riegler2017octnet} and view-based modules
\cite{su15mvcnn,Huang:2017:LMVCNN} for local and global shape processing\ could
be adapted in a similar manner within our architecture. Below we describe the
main focus of our work to learn the parameters of the architecture based on part
hierarchies and tag data.
\vspace{-3mm}
\paragraph{Learning from part hierarchies.}
Our training takes a standard metric learning approach where the parameters
of the PEN are optimized such that pairs originating from the same part sampled from  the hierarchy (positive pairs) have  distance smaller than pairs of points originating from different parts (negative pairs) in the embedded space. Specifically,  given a triplet of points $(a, b, c)$, the loss of the network over this triplet \cite{HofferA14}
is defined as:
\begin{equation}
  \ell(a, b, c) = \big[d(a,b) - d(a,c) + m\big]_+, 
\end{equation}
where $d(a,b) = \norm{\phi_w(a)-\phi_w(b)}_2^2$, $m$ is a scalar margin, and
$[x]_+ = \max(0, x)$.
To avoid degenerate solutions we constrain the embeddings to lie on a
unit hypersphere, i.e., $\norm{\phi(x)}_2^2 = 1$, $\forall x$.
Given a set of triplets ${\cal T}_s$ sampled from each shape $s$ from our dataset $S$, the triplet objective of
the PEN is to minimize the triplet loss:
\begin{equation}\label{eq:tripletloss}
L_{triplet} = \sum_{s\in S} \frac{1}{|{\cal T}_s|} \sum_{(a, b, c) \in {\cal T}_s} \ell(a, b, c).
\end{equation}
\paragraph{Sampling triplets.} One simple strategy to sample triplets is to just access the parts at the finest level of segmentation, then  sample triplets by randomly taking fixed number of similar pairs $(a,b)$
  from the same part  and an equal number of negative points $c$ from another
  part. We call this strategy ``leaf'' triplet sampling.  

An alternative strategy is to consider the part hierarchy tree for triplet sampling. Here, we sample negative point pairs depending on the tree distance between the part groups (tree nodes) they belong to.
Given two nodes $n_i$ and  $n_j$, we use   the sum of path lengths (number of tree edges) from nodes $n_i$ and  $n_j$ to their lowest common ancestor as the tree distance $\delta(n_i, n_j)$  \cite{wiki:LCA} . For example, if the two nodes are siblings (i.e., two parts belonging to the same larger group), then their
lowest common ancestor is their parent and
their tree distance is equal to $2$  (i.e., count two edges that connect them to their parent).
 If two nodes are further away in the hierarchy, then tree distance  increases. In this manner, the tree distance reflects how far two nodes (parts) are in the hierarchy. 

 We compute the probability of 
selecting the positive pair of points from node $n_i$ and the negative pair using the point from
another node $n_j$ as follows:
\begin{equation}
  P(n_i, n_j) \propto \frac{1}{\delta(n_i, n_j)}
\end{equation}
 Sampling points in this way yields more frequent
triplets that consist of negative pairs closer in the hierarchy. Parts that are closer in the hierarchy tend to be
spatially or geometrically closer to each other,
 thus also harder to
discriminate.
We call this sampling strategy as ``hierarchy''  triplet sampling. We
discuss the effect of these two strategies in the experiments
section. 
\vspace{-3mm}
\paragraph{Learning from noisy tag data.}
We can also utilize tag data for segments collected from the COLLADA files, as
described in  Section \ref{sec:dataset}. To train the network using tags, we
add two pointwise fully-connected layers on top of the embedding
network (PEN). 
One way to train
this network is to define a categorical cross entropy
over points whose parts are tagged. However, as shown in Table
\ref{table:dataset-tags}, the total number of tagged points is
small. 
We instead found that a better strategy is to use a one-vs-rest binary
cross entropy loss to also make use of points in un-tagged parts. The
reason is that if a part is not tagged in a shape that has other parts
labeled with tags existing in the shape metadata, then most likely,
that part should not be labeled with any of the existing tags for that
shape (e.g., if a car has tagged parts as `wheel' and `window', then
other un-tagged parts should most likely not be assigned with these
tags). 

More specifically, for every tag in our tag set $\mathcal{L}$ for a shape category, we define a binary cross entropy loss by
considering all points assigned with that tag as `positive' (set $\mathcal{P}$) while the rest of points assigned with other or no tags as `negative' (set $\mathcal{N}$). Given an output probability prediction  for assigning a point $i$ with tag $t$, denoted as $P(y_{i,t}=1)$ produced by the last classification layer (sigmoid layer) of our network, our loss function over tags is defined as follows:
\begin{equation}
L_{tag} \!\! = \! - \! \sum_{t \in \mathcal{T}} \!\bigg( \sum_{i \in \mathcal{P} }\log P(y_{i,t}\!=\!1) + \sum_{i \in \mathcal{N}}\log (1-P(y_{i,t}\!=\!1))  \bigg)
  \end{equation}

\begin{figure*}[ht!]
  \begin{subfigure}{0.5\textwidth}
  \centering
  \includegraphics[width=\linewidth]{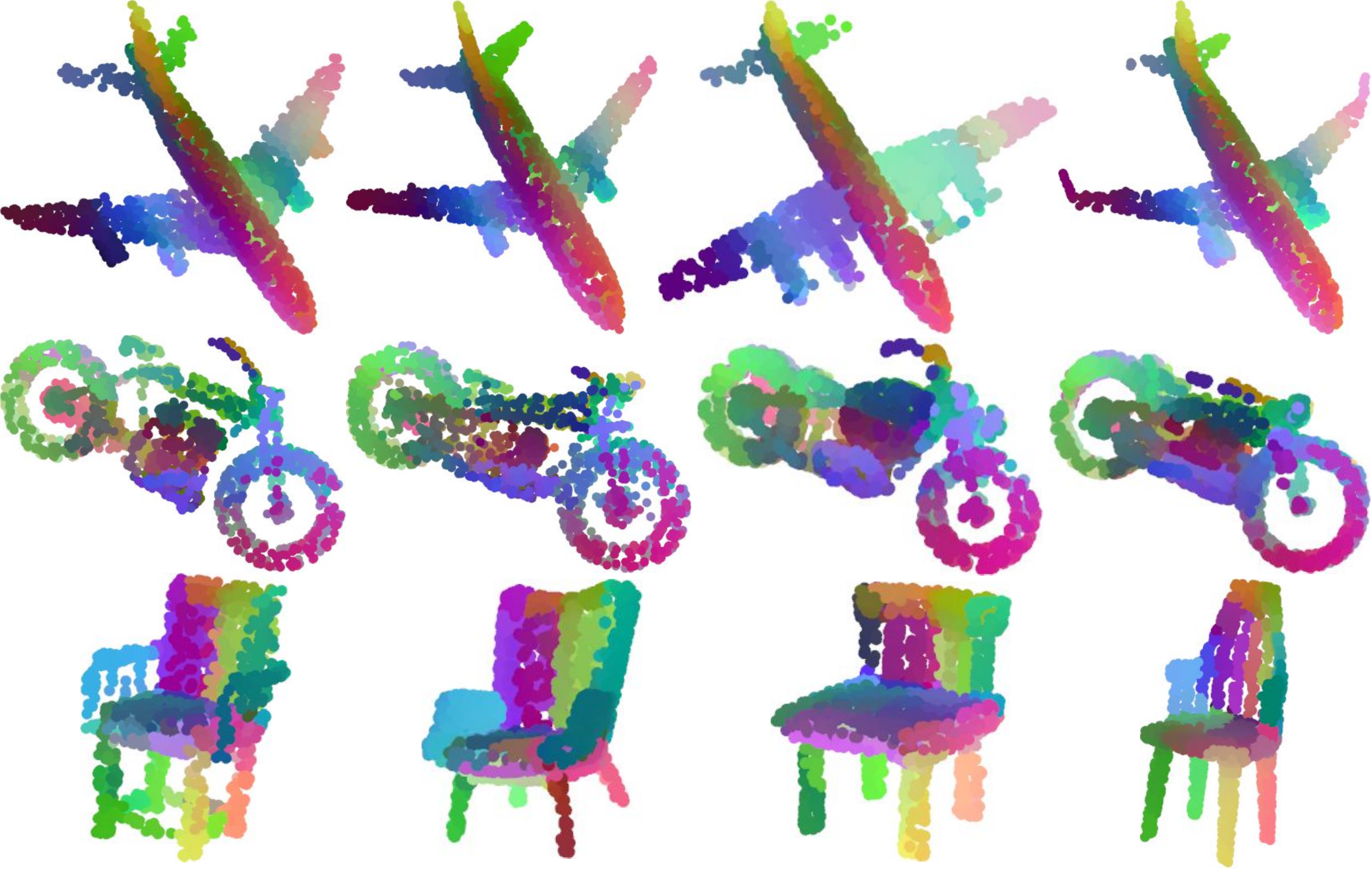}
\end{subfigure}
\begin{subfigure}{0.5\textwidth}
  \label{fig:random-points}
  \centering
  \includegraphics[width=\linewidth]{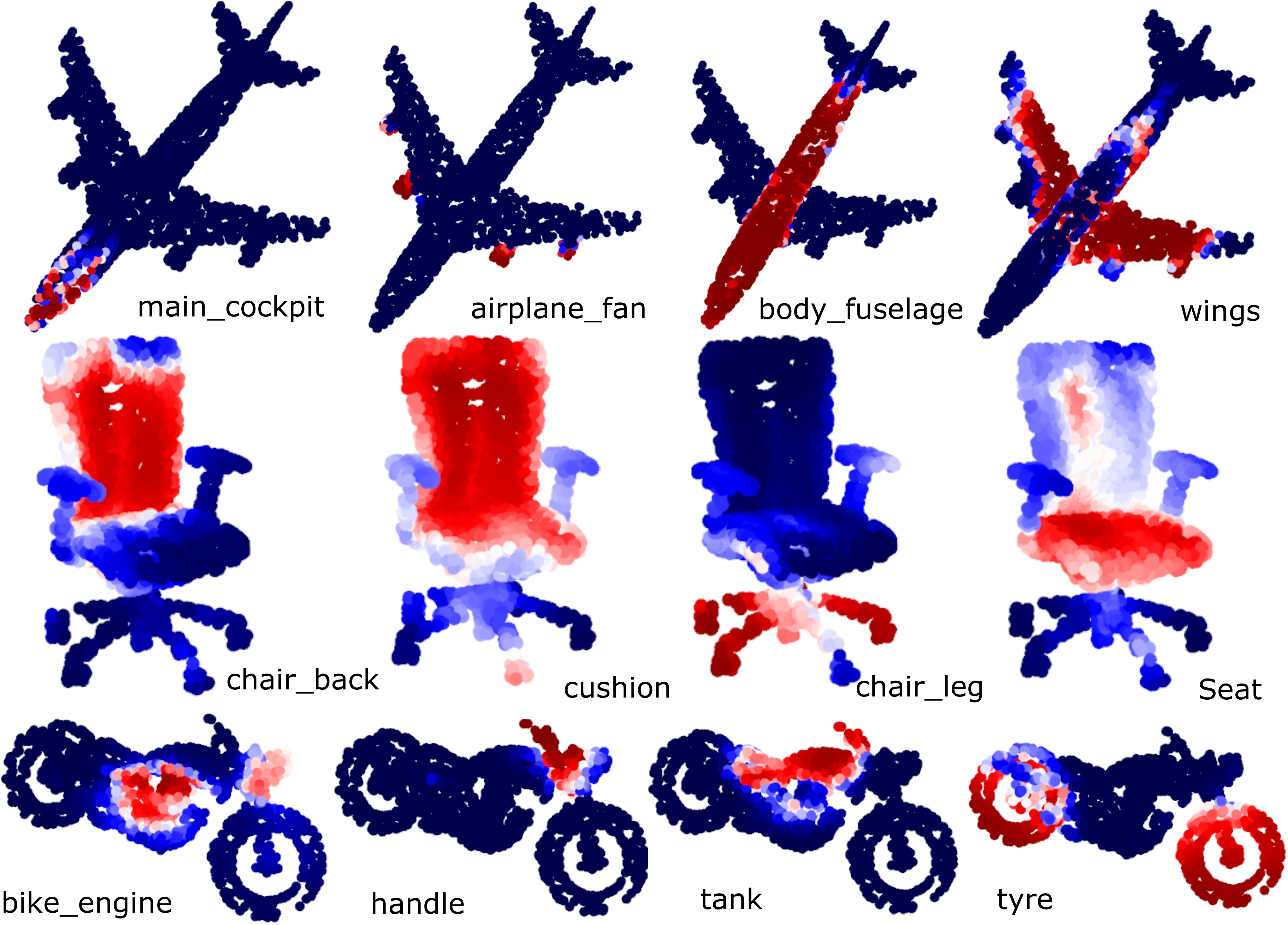}
\end{subfigure}
\vskip -1mm
\caption{\textbf{Visualization of the embeddings.} (Left) T-SNE
  visualization of embedding shown as a color map. Embeddings for similar semantic parts
  are consistently embedded close to each other as reflected by the
  similarity in their color. (Right) Heat map visualization
  of tags predictions across a number of categories and tags. Redder
  values indicate a higher probability of the tag. 
  \emph{(Best seen magnified.)}}
\vskip -2mm
\label{fig:embedding}
\end{figure*}

\paragraph{Training.} We first train our network to minimize the triplet loss
$L_{triplet}$ based on our dataset of shapes that contains part hierarchies.
Training is done in a cross-category manner on $16$ categories\footnote{These
  are the same $16$ categories present in Shapenet semantic segmentation dataset
  from Yi et al. \cite{Yi:2016:SAF}} of ShapenetCore dataset, as described in
Section \ref{sec:dataset}. We use the Adam optimizer \cite{KingmaB14} with
initial learning rate of $0.01$ decayed by the factor of $10$ whenever the
triplet loss stops decreasing over validation set. The mini-batches consist of
$32$ shapes. For further efficiency, in each iteration we randomly sample a
subset of $2.5k$ points (from the $10K$ original points) for each shape during
training. The total number of triplets sampled from a shape is kept constant.

Then for the $5$ categories that include tags, we further fine-tune the learned
embeddings by learning the two additional pointwise fully-connected layer with a
Sigmoid at the end to minimize the tag loss $L_{tag}$. Since tags are distinct
for each category, fine-tuning is done in a category-specific manner (i.e., we
produce a different embedding specialized for each of these $5$ categories).
Although the triplet and tag loss could be combined, we choose a stage-wise
training approach since the shapes with part hierarchies are significantly more
numerous than the shapes that include tags as shown in Table
\ref{table:dataset-tags}. In our experiments we discuss the effect of training
only with the triplet loss, and also the effect of fine-tuning with the tag loss
in each category.

For training networks on few-shot learning task, we do hyper-parameters (batch
size, epochs, regularization etc.) search using validation set of only
one category (`airplane') and use the same hyper-parameters setting to train
all models on all categories in the few-shot learning task.
 \vspace{-3mm}
\paragraph{Few-shot learning.} Given our network pre-trained on our shape
datasets based on part hierarchies and/or tags, we can further train it on
other, much smaller,\ datasets of shapes that include semantic part labels. To
do this, once again we add two point-wise fully-connected layers on top of the
embedding layer, and a softmax layer to produce semantic part label
probabilities. In our experiments, we observe that the part labeling performance
is significantly increased when compared to training our network from scratch
using semantic part labels only as supervision.
 \vspace{-5mm}
\paragraph{Implementation details.}
In our implementation, the encoder of our network extracts $1024$-dimensional
global shape embedding. The decoder concatenates the global embedding with $64d$
point features from encoder, and finally transform it into a $64$-dimensional
point-wise embeddings. Further details of the layers used in PEN are discussed
in the supplementary material. Our implementation is based on PyTorch
\cite{Pytorch}.
\begin{figure*}
  \begin{subfigure}{0.5\textwidth}
  \label{fig:random-shapes}
  \centering
  \includegraphics[width=\linewidth]{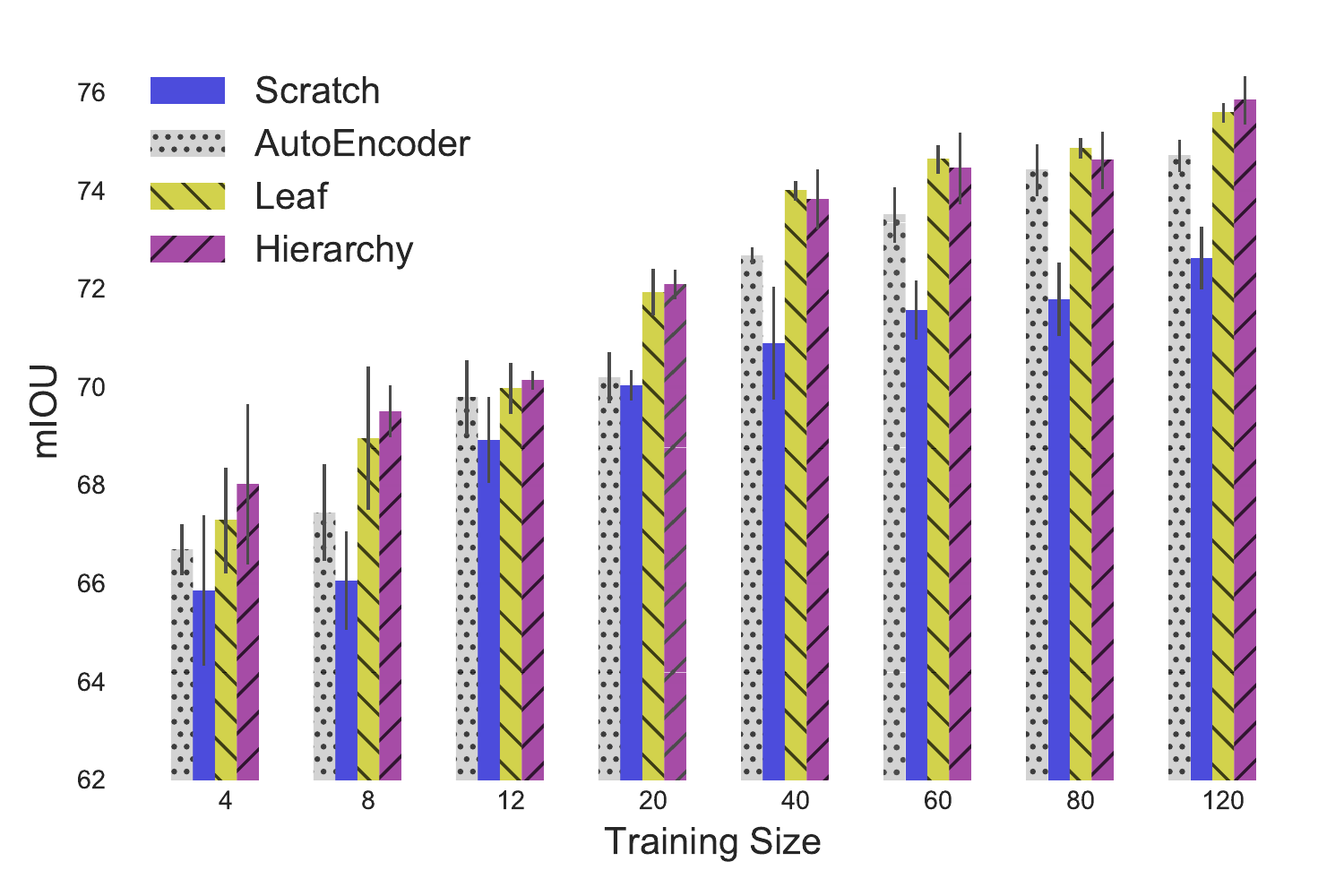}
\end{subfigure}
\begin{subfigure}{0.5\textwidth}
  \label{fig:random-points}
  \centering
  \includegraphics[width=\linewidth]{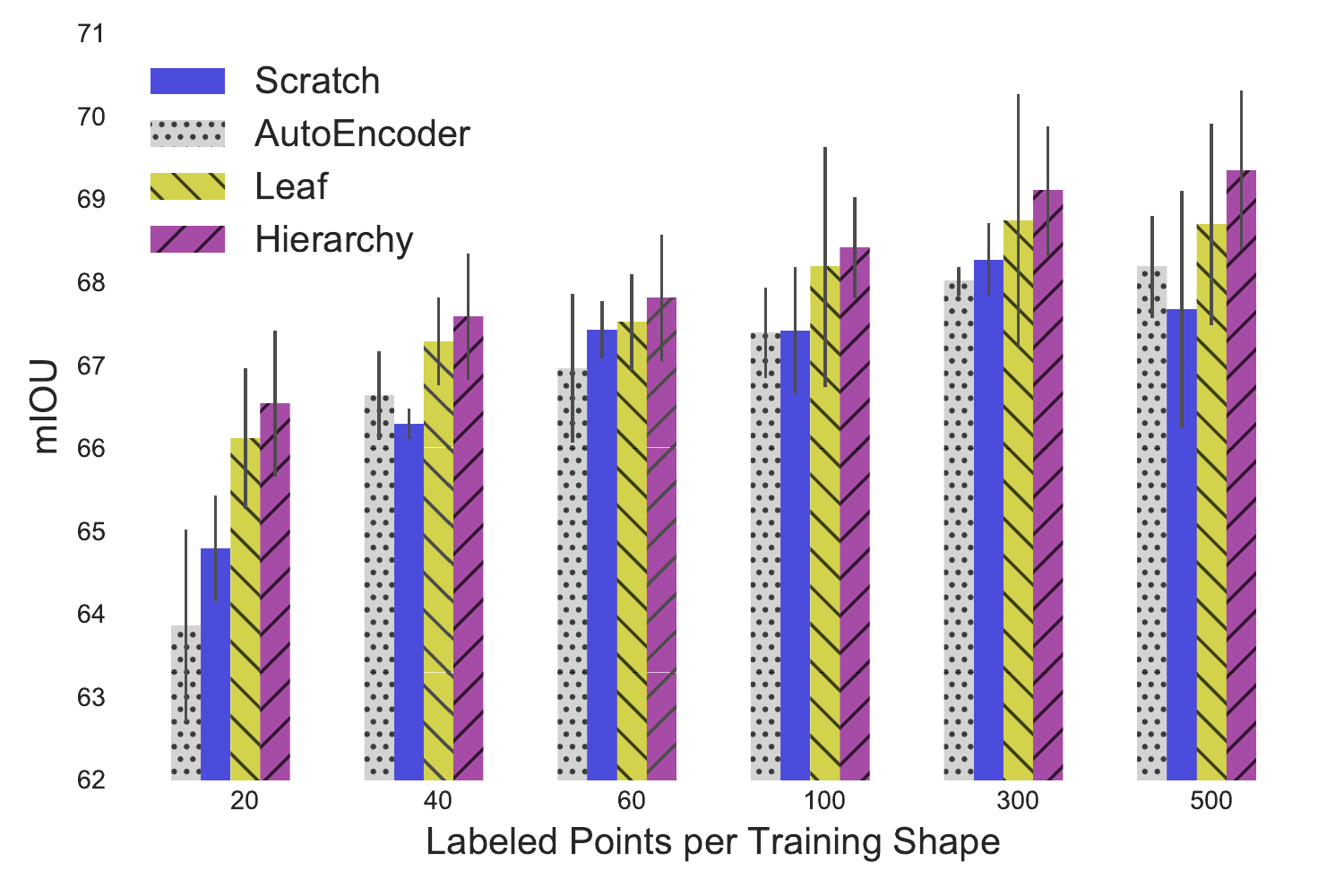}
\end{subfigure}
\vskip -3mm
\caption{\textbf{Benifits of pretraining PEN using metric learning.}
  \textbf{Left:} mIoU evaluation for varying number of training shapes.
  \textbf{Right:} mIoU evaluation for varying number of labeled points and
  fixing the number of training shapes to $8$. We compare different baselines
  and variants of PEN, including training from scratch, autoencoder for
  pre-training, as well as PEN trained with metric learning triplets sampled
  from the leaf of the tree (Leaf) or based on the hierarchy (Hierarchy). PEN
  outperforms both baselines with the hierarchy-based sampling offering a slight
  advantage over the leaf-based one.}
\label{fig:random-shapes}
\end{figure*}
\vspace{-2mm}
\section{Results}\label{sec:results}
\vspace{-2mm}
We now discuss experiments performed to evaluate our method and alternatives.
First, we present qualitative analysis of learned embeddings, then we discuss a
new benchmark we introduce for few-shot segmentation and evaluation metrics, and finally
we present results and comparisons of our network with various baselines.
\vspace{-2mm}
\paragraph{Visualization of the embeddings.} We first present a qualitative
analysis of the PEN embeddings. The embeddings learnt using metric learning only
(without the tag loss) are visualized in Figure \ref{fig:embedding} (left). We use
the t-SNE algorithm to embed the $64$-dimensional point embedding in $3D$ space. Interestingly, the
descriptors produced by PEN consistently embed the points belonging to similar
parts close to each other without explicit semantic supervision. We also
visualize the embeddings predicted by PEN trained with metric learning and
fine-tuned with tag loss in Figure \ref{fig:embedding} (right). The embeddings
have better correspondence with the tags. Interestingly, the network predicts
correct embeddings for points with tags that are not mutually exclusive e.g.
`cushion' and `back' of the chair.\vspace{-4mm}
\paragraph{Few-shot Segmentation Benchmark.}
We anticipate that learning from metadata can improve semantic shape
segmentation tasks, especially in the few-shot learning scenario. To this end we
have created a new benchmark on ShapeNet segmentation dataset
\cite{Yi:2016:SAF}, in which we randomly select $x$ fully labeled examples from
the complete training set for training the network, where $x\in\{4, 8, 12, 20,
40, 60, 120\}$. In this manner, we can test the behaviour of methods with
increasing training number of shapes, starting with the few-shot scenario where
only a handful of shapes (i.e., 4 or 8) is labeled. The performance is measured
as the mean intersection over union (mIOU) across all part labels and shapes in
the test splits. We exclude the shapes existing in our part hierarchy and tag
datasets used for pre-training PEN from the test splits.

We also introduce one more evaluation setting, where for each shape category,
the training shapes have smaller fractions of their original points labeled
($20$, $40$ \ldots $500$) labeled points compared to the
original $2.5K$ points) The case of $\sim20$-$40$ labeled point simulates the
scenario where semantic annotations are collected through sparse user input
(e.g., click few points on shapes and label them).
\begin{figure*}
  \begin{subfigure}{0.5\textwidth}
  \centering
  \includegraphics[width=\linewidth]{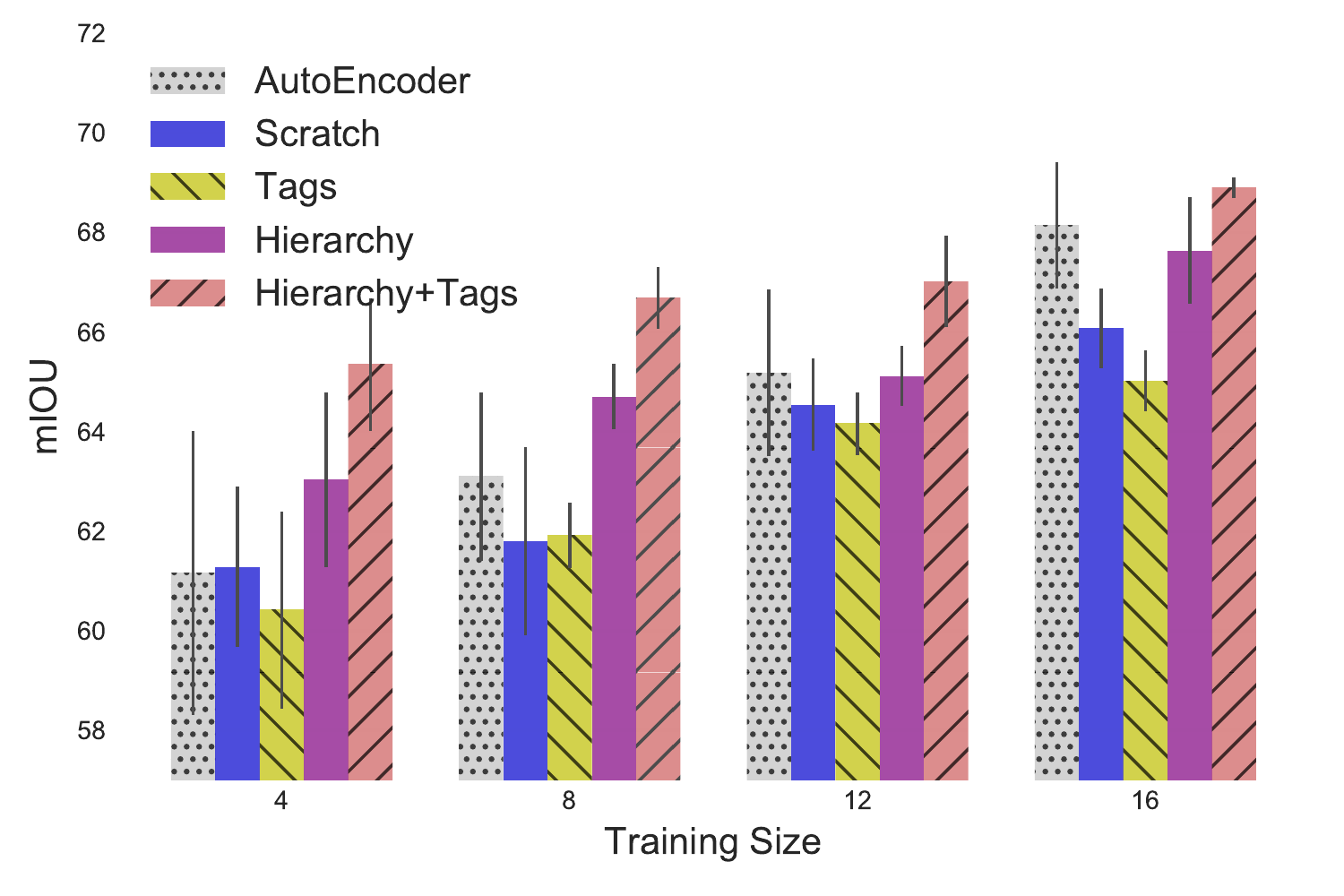}
\end{subfigure}
\begin{subfigure}{0.5\textwidth}
  \label{fig:random-points}
  \centering
  \includegraphics[width=\linewidth]{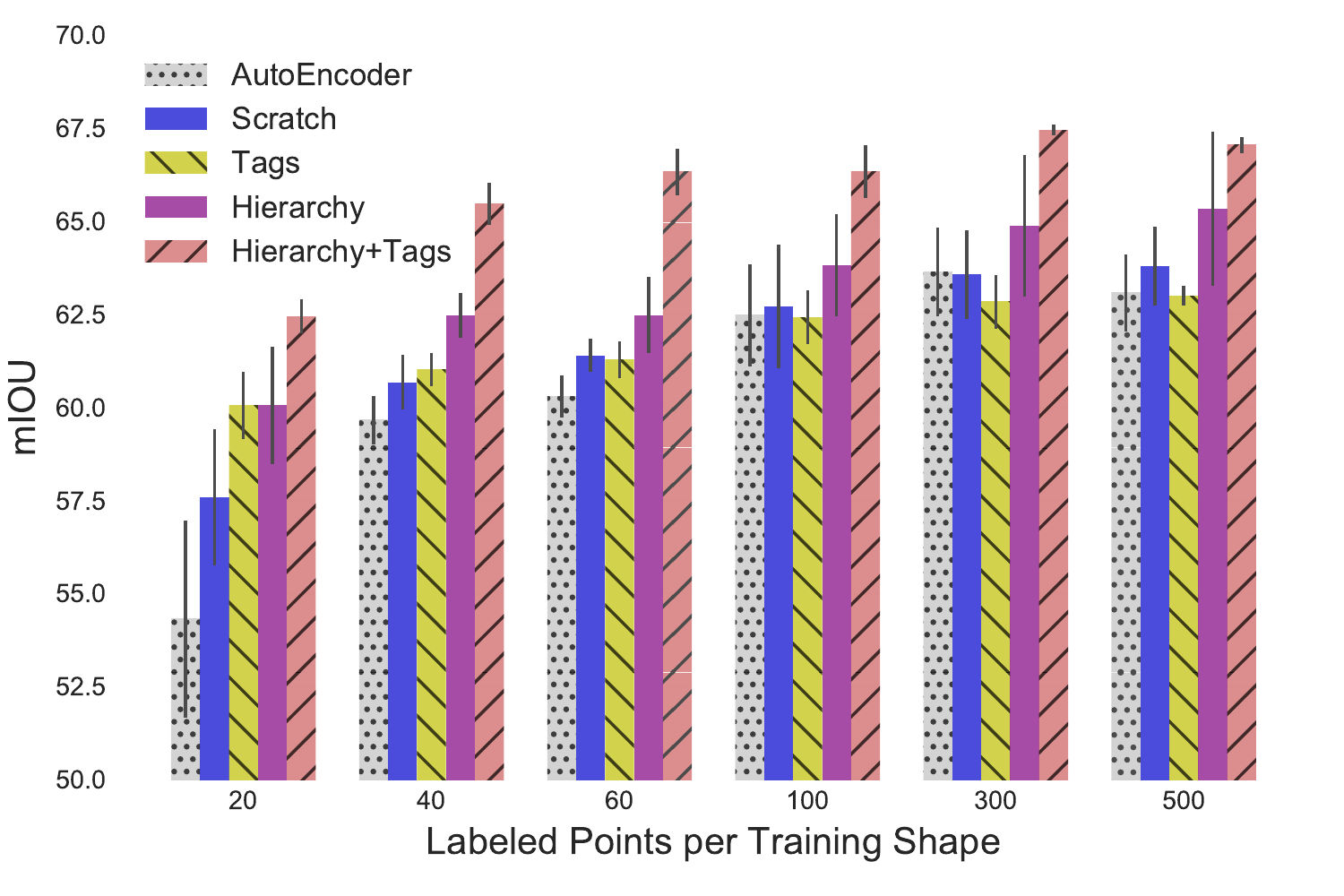}
\end{subfigure}
\vskip -4mm
\caption{\textbf{Benefit of training with tag supervision.} The mIoU
  evaluation when tags are available (5 categories:\ motorcycle,
  airplane, table, chair, car). We include the same baselines and PEN 
  variants as Figure \ref{fig:random-shapes}, including two more
  PEN variants: one trained with tags only (Tags) and another trained
  both on hierarchy and tags (Hierarchy + Tags). 
  \textbf{Left:} Shows the performance in the few-shot
  setting. \textbf{Right:} Shows the performance in the few-point
  setting. In both cases the tag data (Hierarchy + Tags) provides additional benefits
  over the PEN models trained with the hierarchy supervision
  (Hierarchy). Tag data alone is not as effective as the autoencoder
  since the supervision is highly sparse.}
\label{fig:tags-random-shapes}
\end{figure*}
\vspace{-4mm}
\paragraph{Baselines.}
Since we utilize a vast number of unlabeled data from the same domain it is
important to compare with baselines. Our first baseline simply trains PEN from
scratch on the training splits of our few-shot segmentation benchmark using only
semantic label supervision (without using metadata). Second, we also compare
with another baseline, where we train an autoencoder network that leverages
only geometry as an alternative to produce point-wise embeddings. This network
first encodes the input point cloud to point-wise embeddings producing a
$1024$-dimensional point-wise representations exactly as in PEN, then a decoder
uses upconvolution to reconstruct the original point cloud. The Chamfer distance
between generated points and input points is used as a loss function to train
this network. We first pre-train the autoencoder on the shapes included in our
part hierarchy dataset. After this pre-training step, we retain the encoder and
replace the geometry decoder with PEN's decoder and add two pointwise fully
connected layers and a classification layer to produce semantic part label
probabilities. The resulting network is then trained in stages, first the
decoder and then the entire network at smaller learning rate, on the training
splits of our few-shot segmentation benchmark.

Finally, we also evaluate the two strategies to pretrain the embedding network
using different triplet sampling techniques \ie leaf-level shape parts (``leaf''
triplet sampling) and based on using the hierarchy tree (``hierarchy'' triplet
sampling) as described in (Section \ref{sec:method}).

Next, we compare the performance of our method with the baselines and different
sampling strategies under the scenario of using only the triplet loss and
cross-category training. Then, we discuss the performance in the case where we
additionally use the tag loss. \vspace{-3mm}
\paragraph{Few-shot Segmentation Evaluation.}
In Figure \ref{fig:random-shapes} (left), we plot the mIOU of the baselines
along with our method. The plotted mIOU is obtained by taking the average of the
mIOU on our test splits over all categories and repeating each experiment $5$
times. The network trained from scratch (without any pre-training) has the worst
performance. The network based on the pre-trained autoencoder shows some
improvement since its point-wise representations reflect local and global
geometric structure for the point cloud reconstruction, which can be also
relevant to the segmentation task. Our method consistently outperforms the
baselines. In particular, the ``hierarchy'' triplet sampling that uses the part
hierarchy trees to choose triplets for training our network performs the best on
average. A $3.5\%$ mIOU improvement ($10.2\%$ drop in relative error) is
observed compared to training from scratch at $8$ training examples -
interestingly, the improvement is retained even for $120$ training examples. The
``hierarchy'' triplet sampling also improves over the ``leaf'' triplet sampling
until $20$ training examples, then their difference gap between these two
strategies is closed.
\vspace{-3mm}
\paragraph{Evaluating with limited labeled points per shape.}
In the previous section we observed the performance of our method and baselines
by changing the number of training shapes. Here we also examine the performance
in the few-shot setting where we keep the number of training shapes fixed and
vary the number of labeled points per training shape. We retrain the above
baselines (train from scratch, autoencoder)\ and triplet sampling strategies
(``leaf'' and ``hierarchy'') with $8$ training examples, and vary the number of
labeled points as shown in the Figure \ref{fig:random-shapes} (right). Again our
network using the ``hierarchy'' triplet sampling performs better than the
baselines. It offers $1.7\%$ better mIOU ($4.9\%$ drop in relative error)
compared to training from the scratch using $20$ labeled points.
\vspace{-3mm}
\paragraph{Are tags useful?}
Here we repeat the two few-shot sementation tasks on 5 shape categories
(motorcycle, airplane, table, chair, car) that include some tagged parts in
their shape metadata. Here, we examine two more PEN variants: (a) PEN
pre-trained using the tag loss only (no triplet loss), then fine-tuned on the
training splits of our semantic segmentation benchmark (this baseline is simply
called ``tags'' network), 2) our network pre-trained using triplets loss based
on the ``hierarchy'' sampling, then fine-tuned with the tag loss, and finally
further fine-tuned on the training splits of our semantic segmentation benchmark
(this baseline is called ``Hierarchy+Tags'' network). The two PEN variants are
trained per each category of the $5$\ categories. The results are shown in
Figure \ref{fig:tags-random-shapes}.

When using $8$ training examples, the Hierarchy+Tags network offers $4.8\%$
better mIOU ($12.8\%$ drop in relative error) on average compared to training
from scratch in these 5 categories (refer Figure \ref{fig:tags-random-shapes}
(left)). An improvement of $2.8\%$ mIOU ($8.3\%$ drop in relative error) is
maintained for $16$ training examples. Similarly, when using $20$ labeled points
per shape, Hierarchy+Tags performs $4.9\%$ mIOU better ($11.47\%$ drop in
relative error) than training from scratch (refer Figure
\ref{fig:tags-random-shapes} (right)). In general, the Hierarchy+Tags PEN\
variant outperforms all other baselines (training from scratch, autoencoder)
and also the variant pre-trained using tags only (``Tags'' network) on both
evaluation settings with limited number of training shapes and limited number of
training points. This shows that the combination of pre-training through metric
learning on part hierarchies and fine-tuning using tags results in a better,
warm starting model for semantic segmentation task.

\vspace{-2mm}
\section{Conclusion}
\vspace{-1mm}
We presented a method to exploit existing part hierarchies and tag
metadata associated with 3D shapes found in online repositories to
pre-train deep networks for shape segmentation.
The trained network can be used to ``warm start'' a model for semantic
shape segmentation, improving performance in the few-shot setting. 
Future directions include investigating alternative architectures and
combining other types of metadata, such as geometric alignment or
material information.

\textbf{Acknowledgements.} This research is funded by NSF (CHS-161733
and IIS-1749833). 
Our experiments were performed in the UMass GPU cluster obtained under
the Collaborative Fund managed by the Massachusetts Technology
Collaborative.

{\small
  \bibliographystyle{ieee_fullname}
\bibliography{egbib}}
\end{document}